THE PRODUCTION OF PROBABILISTIC ENTROPY

IN STRUCTURE/ACTION CONTINGENCY RELATIONS [*]

Loet Leydesdorff

Department of Science and Technology Dynamics

Nieuwe Achtergracht 166

1018 WV   AMSTERDAM

The Netherlands

December 1993

Abstract

Luhmann (1984) defined society as a communication system which is structurally coupled to, but not an aggregate of, human action systems. The communication system is then considered as self-organizing ("autopoietic"), as are human actors. Communication systems can be studied by using Shannon's (1948) mathematical theory of communication. The update of a network by action at one of the local nodes is then a well-known problem in artificial intelligence (Pearl 1988). By combining these various theories, a general algorithm for probabilistic structure/action contingency can be derived. The consequences of this contingency for each system, its consequences for their further histories, and the stabilization on each side by counterbalancing mechanisms are discussed, in both mathematical and theoretical terms. An empirical example is elaborated.

THE PRODUCTION OF PROBABILISTIC ENTROPY

IN STRUCTURE/ACTION CONTINGENCY RELATIONS

The emphasis on "self-referentiality" and "autopoiesis" in sociological theory enables us to use models from evolutionary biology for the analysis of social systems and their developments. Additionally, Luhmann (1984) proposed that society can be considered as constituted not of human beings, but of communications. This makes the study of social phenomena accessible for the mathematical modelling of communication patterns by means of Shannon's mathematical theory of communication (Shannon 1948). Luhmann (1984; 1990), however, elaborated this empirical perspective only with qualitative reference to evolution theory.

Society as a system of communications

Of course, communications are generated by human beings and have to be understood by human beings, but Luhmann's sociology defines society as the network which is *added to* the actors as the nodes. The network differs in nature and in operation from the individual systems: individual "consciousness systems" process thoughts on the basis of perceptions; while society processes communications. The two systems are coupled *structurally*, i.e., they presuppose each other in the operation, but the one is not an aggregate of the other. In addition to the four billion or so people who perform their own self-referential loops (e.g., "thinking"), society is also a system. This system communicates with the actors at the nodes in terms of co-variances, while it exhibits auto-covariance in the remaining variance during any discrete time period. The social system is therefore self-referential. (Whether self-referential social systems are also self-reproducing and reflexive, and therefore "autopoietic," remains another empirical question (cf. Teubner 1988; Leydesdorff 1993b).)

In modern societies, communications are functionally differentiated. For example, one can communicate through market-transactions in the economy, or through love in personal relations. On the one hand, the focus on communications makes the theory of symbolic generalized communication media (cf. Parsons) the starting point for the development of the special sociologies (Luhmann 1982, 1988, and 1990). On the other hand, the reformulation of the unit of operation with reference to the communication system--instead of exclusive reference to the discrete actors--bridges the gap with symbolic interactionism as the other great tradition in American sociology (Luhmann 1975; Leydesdorff 1993c).

When the system is functionally differentiated, the structural coupling between actors and the social communication system may take different forms in the various subsystems. However, these forms are functionally equivalent, and thus we may expect that the structure/action contingency relation can be studied in terms of one underlying general algorithm (cf. Giddens 1979; Burt 1982).

<u>The mathematical theory of communication</u>

Shannon (1948) deliberately chose to define information so that it would correspond with Boltzmann's (1877) definition of entropy in thermodynamics. In theoretical biology, evolutionary processes are increasingly studied in terms of non-equilibrium thermodynamics, i.e., as entropy generating processes (e.g., Brooks and Wiley 1986). Although Shannon's definitions have been controversial in the context of thermodynamics (e.g., Brillouin 1962; Wicken 1987), they have been less so when applied to social phenomena (e.g., Georgescu-Roegen 1971; Theil 1972;

Krippendorf 1986). The social sciences analyze variances: the variance of a distribution is expected to contain information which is part of the uncertainty in the system which exhibits the variation. This uncertainty can be considered as probabilistic entropy (cf. Bailey 1990).

At the technical level, it is possible to develop a continuous version of information theory using integral calculus (e.g., Shannon 1948; Theil 1978). However, for social phenomena the original, discrete version (which is also mathematically simpler) is more adequate, since actions are discrete instances. Measures of information theory in this discrete form are composed of sigmas, which allow for the systematic study of the processes of aggregation and disaggregation, and for the effects of groupings. Additionally, since the measures are non-parametrical and based only on probabilities (i.e., relative frequencies), neither the dimensionality of the problem nor the measurement scale is inhibitive (see also: Krippendorff 1986). Information theory can be developed in order to integrate static and dynamic forms of analysis (Theil 1972; Leydesdorff 1991).

Let me now introduce the most important formulas. If we define $h$ as the information content of the message that an event has occurred, then the expected information content of the distribution of a variable with relative frequency $p_i$ can be written as:

$$H = \Sigma_i \, p_i * h_i \quad (1)$$

By using Shannon's (1948) classical function for information ($h_i = {}^2\log(1/p_i)$),[i] we may write:

$$H = -\Sigma_i \, p_i \log p_i \quad (2)$$

and for the multi-variate case:

$$H = -\Sigma_i \Sigma_j \Sigma_k\ p_{ijk} \log p_{ijk} \qquad (3)$$

As with chi-square, *H* can be used as a measure of the association among variables. The overall uncertainty for two variables *x* and *y*, *H(x,y)*, is equal to *H(y)* plus the amount of uncertainty which *x* adds to it, given the uncertainty in *y*, i.e., *H(x|y)*. (See <u>Figure 1</u> from (Attnaeve 1959) for a visual representation.)

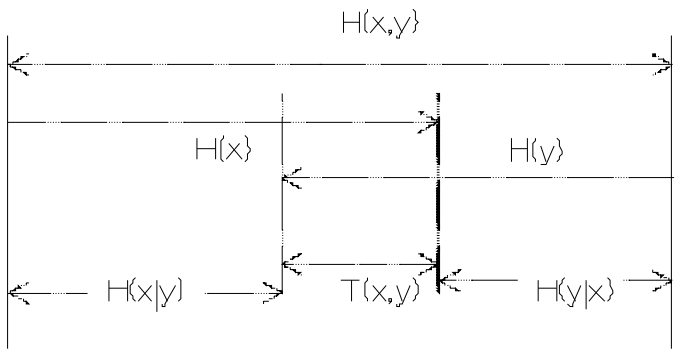

Therefore:

$$H(x,y) = H(y) + H(x|y) \qquad (4)$$

The *mutual information* or *transmission* between *x* and *y* is consequently defined as:

$$T(x,y) = H(x) - H(x|y) = H(y) - H(y|x) \qquad (5)$$

This is the reduction in the uncertainty of the prediction of *x*, given knowledge about the distribution of *y*.

While "delta chi-square" does not have a clear interpretation,[ii] the decomposition of $H$ (and $I$, below) in terms of the contribution to the uncertainty of each of the component cells (or subsets) is straightforward.[iii] Additionally the following formula can be derived for the disaggregation of $H$ into $g$ groups (Theil, 1972):

$$H = H_0 + \Sigma_g P_g * H_g \qquad (6)$$

$H_0$ is a measure of the uncertainty among the groups $g$, or in other words, a measure of the specificity of the distribution of the relevant variables within the groups.

On the basis of the above definition of information, it can be shown (see, e.g., Theil 1972) that if we have a system of mutually exclusive events, $E_i$, with prior probabilities $p_i$, then the expected information content $I$ of the message which transforms the *prior* probabilities $p_i$ into the *posterior* probabilities $q_i$ is given by the following expression:

$$I = \Sigma_i \; q_i \; * \; \log(q_i / p_i) \qquad (7)$$

Correspondingly, for the multi-variate case, the expected information content of the message transforming the prior probability distribution $p_{ijk}$ of events into the posterior probability distribution $q_{ijk}$, is equal to:

$$I = \Sigma_i \Sigma_j \Sigma_k \; q_{ijk} \; * \; \log(q_{ijk} / p_{ijk}) \qquad (8)$$

Although overall $I >= 0$,[iv] $\wedge\ I$ can become negative for a term if $q < p$.[v] (Of course, $\wedge\ H$ in formula (2) is always $>= 0$.) Furthermore, in the dynamic case the grouping rules among levels of aggregation are somewhat more complex than in the static case (cf. Theil 1972).

However, in principle, the two formulas, i.e., for *H* and *I*, provide us with a complete framework for the development of a set of methodologies equivalent to multi-variate analysis and to time series analysis, respectively (Leydesdorff 1991).[vi]

The structure/action contingency relation

Let *A* be a structure and *B* a distribution of actors such that at each moment in time the actors will take action *given* this structure; thus, *B* is conditioned by *A* when it operates. Action which can thus be defined as *B*/*A*, has an impact on structure *A* only in instances thereafter. (See Figure 2 for a visual representation.)

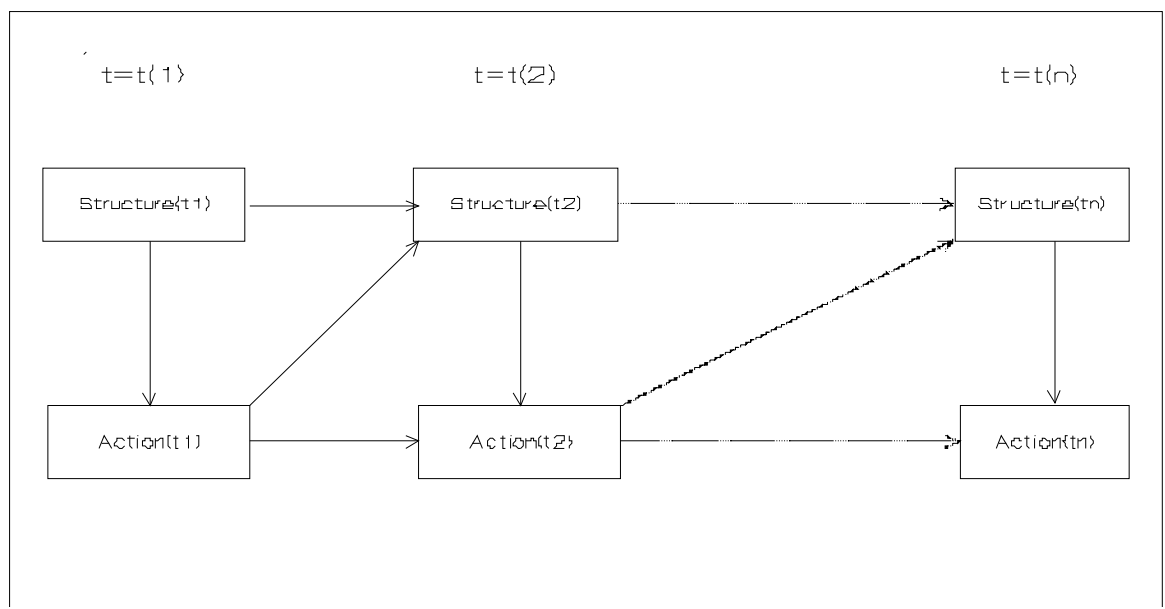

Structure (*A*) can also be considered as a network of which these actors (*B*) are the nodes (cf. Burt 1982). At each moment in time, the network conditions all the actors; if action takes place at any node(s), this conditions the network at the next moment.[vii] This model can be easily generalized to the model of parallel distributed computers (see also: Pearl 1988; Leydesdorff 1993c): each actor is comparable to a local processor which acts on the basis of its own programme given the conditions set by the network; while the operation of the network, which operates according to its own programme, is conditioned by the sum total of the previous actions.

Let us now describe these systems in information-theoretical terms. Structure *A*, which is expected to contain $H(A)$ at time *t* conditions at that moment action(s) at *B*. Action can therefore be expected to contain only $H(B|A)_t$ information. At the next moment ($t + 1$), structure *A* is informed by the message that action(s) at *B* has/have taken place, and therefore the system thereafter contains an uncertainty $H(A|B)_{t+1}$. In a later section I decompose this *a posteriori* information content in the information contents of the *a priori* systems (*A* and *B*/*A*, respectively), but let me here first focus on the dynamics of the message that action has occurred.

By using equation (7), we may write the information content of this message as follows:

$$I_{(A|B \, : \, A)} = \Sigma \; q(A|B) * \log\{ \; q(A|B) \; / \; p(A)\} \qquad (9)$$

in which $\Sigma \; q(A|B)$ represents the *a posteriori* state of the structure (*A*), i.e., after action (*B*) has occurred, while $p(A)$ describes the *a priori* probability distribution. In other words: this formula expresses the information content of the message that the structure has self-referentially to update its information content since actions at the network nodes *B* have occurred.

Since, according to the third law of the probability calculus:

$$p(A \text{ and } B) = p(A) * p(B|A) \qquad (10)$$

$$= p(B) * p(A|B) \qquad (10')$$

we may also write:

$$p(A|B) = \frac{p(A) * p(B|A)}{p(B)} \qquad (11)$$

This is known as Bayes' Formula, and expresses the *a posteriori* probability as a function of the *a priori* ones. Indices for *a priori* (*t*) and *a posteriori* (*t* + 1) probabilities can be written as follows:

$$p(A|B)_{t+1} = \frac{p(A)_t * p(B|A)_t}{p(B)_t} \qquad (11')$$

or equivalently, and using *q* as indicator of the *a posteriori* probability:

$$p(A) = \frac{q(A|B) * p(B)}{p(B|A)} \qquad (12)$$

Substituting (12) into (9) one obtains:

$$I(A|B_{t+1} : A_t) = \Sigma\ q(A|B) * \log\{\ p(B|A)\ /\ p(B)\} \qquad (13)$$

This formula can also be written as the difference between two logarithms:

$$I_{(A|B : A)} = \Sigma\ q(A|B) * \log\{\ q(A|B)\ /\ p(B)\} +$$
$$-\ \Sigma\ q(A|B) * \log\{\ q(A|B)\ /\ p(B|A)$$
$$= I_{(A|B : B)} - I_{(A|B : B|A)} \quad (14)$$

This difference is equal to an improvement of the prediction of the *a posteriori* structure: the expected information content of the message that A is conditioned by B (left-hand side of the equation) is equal to an *improvement* of the prediction of the *a posteriori* distribution ($\Sigma\ q(A|B)$) if one add to one's knowledge of the *a priori* distribution ($\Sigma\ p(B)$) the information about how the latter distribution was conditioned by the network distribution ($\Sigma\ p(B|A)$).

The crucial point is the implied shift in the systems of reference in formula (13). (This possibility is an analytical consequence of the dynamic, and therefore non-trivial interpretation of Bayes' Formula above.) The right-hand factor of the right-hand term, i.e., ($\Sigma\ p(B|A)\ /\ p(B)$) in formula (13), describes the instantaneous conditioning of *action* by structure at time *t*, while the left-hand factor refers to the description of the *network* after action, i.e., $\Sigma\ q(A|B)$ at the next moment. Therefore, the formula explicates how action at the nodes and the network are conditioned mutually and dynamically.

In other words: if one initially (at time *t*) had knowledge only of the distribution of the nodes, i.e., the actors ($\Sigma\ p(B)$), and then became informed of how, at this moment, action at the nodes is distributed given the network ($\Sigma\ p(B|A)$), formulas (9) and (13) teach us that this provides us with the same expected information (*I*) about the network distribution, given the nodes at *t* + 1 (i.e., the *a posteriori* distribution), as does the message that the network distribution is conditioned by the nodes. The improvement is equal to the prediction based on the *a priori* network. Thus, one can study the coupled systems from either side, but one learns only in terms of their interaction.

Interaction, conditionalization and the "freedom" of each system

Let us first focus on the interpretation of the *a priori* system which can be described with $\Sigma\ p_{(B|A)}/p_{(B)}$ (in the right-hand factor of formula (13)). Obviously, this is a ratio; the denominator can be considered as a normalization term for the size of the action system in terms of the number of actors involved.[viii]

If the actions are independent $p_{(B)} = \Sigma_A\ p_{(B|A)}$, and in this case:

$$p_{(B|A)}/p_{(B)} = p_{(B|A)}\ /\ \Sigma_A\ p_{(B|A)} \qquad (15)$$

In qualitative terms this normalization means that a *structure A* as a self-referential system can only incorporate information in terms which have been conditioned by itself, and which have been normalized. For example, in a democracy as a normative system, the independence of votes is assumed, and the system accepts a decision with 51% in favour, regardless of whether the total number of voters in the system was a hundred or a million. Secondly, for example, a theoretical system like Newtonian mechanics cannot accept the upward motion of a plume as counter-evidence to the laws of gravity, since the only relevant action is to produce evidence which can be made available to it in its own theoretical terms. In the case of a theoretical system, this normalization obviously refers also to the statistics.

Note that the normalization is symmetrical in *A* and *B*:

$$p_{(B|A)} / p_{(B)} = \{p_{(B|A)} * p_{(A)}\} / \{p_{(B)} * p_{(A)}\} = p_{(A \text{ and } B)} / p_{(A)} * p_{(B)} \quad (16)$$

What has been said of structure, is thus also true of action insofar as action is to be considered as a system. If actions, however, are not independent in a system of actions, equality (15) no longer holds.

Actions can be coupled in a system (the actor), and this system can be "autopoietic" or not, dependent upon the form of the coupling. In other words: there are various ways in which actions can be aggregated. An action system is autopoietic if the actor is an individual who relates the actions internally. For example, if a scientist first must do research and only then can publish it, there is additional uncertainty created within his/her system since there are various ways in which the two acts can be related. Analogously, if there are (feedback) relations among actions by different actors in an organization (e.g., in a research community or a department), this system can have a more systematic impact on the relevant structures (e.g., the scientific communication system), since the assumption of independence among the actions is no longer valid. I return to the issue of the generation of uncertainty and redundancy by other actors in a later section.

In summary, a self-referential structure does not merge with the uncertainty in the events within its environment, but only with the uncertainty which it has previously conditioned within its environment, and after normalization for the size of the events. In other words: structure and action exchange information only insofar as they are coupled, and only to the extent that the other system is performing above or below the *a priori* expectation.

However, how *B* is conditioned by *A* does *not* inform us about how *A* is conditioned by *B*, but only about their "mutual information," i.e., the extent to which the conditioning *reduces* the remaining uncertainty. Knowledge of this (static!) transmission between action and structure reduces the uncertainty in the transmission at the next moment. The remaining uncertainty in *A* (or *B*) remains always underdetermined (i.e., "free"): each system contains its own total uncertainty, on the basis of which it enters self-referential loops in which it conditions the actions which generate relevant information for it. After the cycle (*a posteriori*) the fact that a specific interaction with *B* has occurred belongs to the history of *A* (i.e., "is a given for *A*"). I shall show in a later section that this necessarily adds to the information content of *A* (i.e., by making history, *A* increases its probabilistic entropy). However, the system's uncertainty can increase only to the maximal amount of uncertainty it can contain, which is equal to the logarithm of its elements (log $n_A$).

The freedom of the two systems with respect to each other can also be seen mathematically on the basis of another interpretation of the (above noted) quotient between $p_{(B|A)}$ and $p_{(B)}$. As noted the two factors refer to "actions" and "actors", respectively. The difference between the two factors ($p_{(B)}$ - $p_{(B|A)}$) corresponds with that part of the uncertainty in *B* which is determined by *A*. At any moment, knowledge of the uncertainty in the network improves our prediction of the uncertainty at the nodes of the network, but only for the part of the transmission, i.e., $H_{(B)} - H_{(B|A)}$ ( = $T_{(AB)}$) (see also Figure 1). It informs us only about the static transmission at that moment, i.e., about the impact of the vertical arrow in Figure 2, which indicates the instantaneous conditioning of actors by structure (cf. Giddens 1979). However, the other part of the uncertainty ($H_{B|A}$) remains precisely undetermined by this conditioning, since it is only conditioned by it. *Mutatis mutandis* the same holds true for $H_{(A|B)}$.

Apart from the question of whether the actions are independent (i.e., whether equation (15) holds), the general algorithm for action/structure contingency relations (equation (13)) is necessarily true. The operation of the action system (e.g., actors) generates new information for structure, but the information adds to that part of the uncertainty in structure which is determined by this action system, and not to the remaining uncertainty. Therefore, left alone, two coupled systems would increasingly co-determine each other. However, in a multi-actor system other actions can add information to structure meanwhile, and thereby change the conditionalization. The synchronization of the relevant actions by structure leads to Markov chain transitions as a special case.

In summary: action pre-sorts information for structure, while it is itself conditioned by structure. However, the total information content of structure is independent of action: each system remains "free" at each moment in time. The improvement in the prediction is based only on the local interaction. Furthermore, because of the sigma in the algorithm the improvements are additive, and can therefore also be decomposed into single actions or subgroups of actions.

Empirical example

As noted, the resulting improvement in the prediction of *A* at the later moment is analytically due to the mutual information between *A* and *B* at the earlier moment. It is an evaluation of how much our knowledge of this transmission at the earlier moment informs us about the conditioning at the later moment. If there is conditionality in the (static) probabilistic relations between *A* and *B*, there is necessarily also dynamic coupling.

If the systems *A* and *B* are completely coupled in one operation, the $I_{(A|B\,:\,A)}$ is equal to $I_{(B|A\,:\,B)}$, since the relations are symmetrical in *A* and *B*. However, if each of two systems pursues its own respective operation, the effects of their boundedness by each other will be asymmetrical. Since *I* can also be used as a measure of the quality of the prediction, the formulas allow us to develop a measure of whether (and to what extent) the one type of data in empirical research on structure/action contingencies represents structure, while the other represents action, or vice versa. One can also easily imagine designs in which what is action at one moment in time will operate as structure at a later moment. Methodologically the two perspectives provide symmetrical tests, just as they do conceptually in the idea of a "mutual shaping" of structure and action by one another.

For an empirical illustration, let me use the transaction matrix of the aggregated citation data of 13 major chemistry journals among each other. These matrices can be easily compiled using the <u>Journal Citation Reports</u> of the <u>Science Citation Index</u>.[ix] I shall use the 1984 matrix for the journals listed in Table 1 as the *a priori* distribution, and the 1985 one as the *a posteriori* distribution.

In each year, the matrix contains a "being cited pattern" that can be taken as structure, and the "citing" side can be considered as action. By citing one another the journals reproduce structure in a subsequent year. Since the citation matrix contains information with respect to both the cited *and* the citing dimension, it should provide us with an opportunity to make predictions about this reproduction of structure.

In 1984, we can compute the (static) transmission between the "cited" and the "citing" side of this matrix by using formula (5). The expected information content of this message is 964.17 mbits of information. For 1985, this mutual information is 972.48 mbits, i.e., 8.31 mbits more. However, in the dynamic model (i.e., by using formula (13)) we find an improvement in the prediction for 1985 to 969.73 mbits on the basis of 1984 data.[x] This means that 5.56 mbits of the 8.31 mbits change in the transmission (or 66.9%) can be attributed to the previous transmission. In other words: the increase in the coupling is above expectation. (One reason for this may be that there occurred, for example, a grouping among the cited into a structure or feedback among the citing journals.)

Since the operations of "cited" and "citing" are mutual in this universe of 13 journals, this result remains the same when the matrix is transposed. The two systems are completely coupled, since there is only one operation, *viz.* citation. However, in the parallel and distributed computer model, one may also assume a communication system between the cited and the citing journals, in which the operations are mediated by the network, and thus in principle asymmetrical. In this case, one needs an independent operationalization of structure in the communication system, e.g., in terms of the eigenstructure of the matrix. Note that the eigenstructure of an asymmetrical matrix is asymmetrical indeed.

In order to keep the analysis simple, let us make the (reasonable)[xi] assumption that this set of 13 journals can be grouped in three sets, namely: one of inorganic chemistry journals, one of organic chemistry journals, and one of journals which belong to the specialties of physical chemistry and chemical physics. (The attribution of the journals to these groups is given in Table 1.) This provides us, in a second research design, with a matrix of three cited clusters which represent the cited structure, and 13 citing journals which represent action. At a third level, we may then also group the citing journals, and analyze the three by three matrix which represents the interaction between the presumed cited and citing structures.

As above, on the basis of the 1984 matrix we can make a prediction of the transmission in the 1985 matrix. In the case of the asymmetrical matrix of three cited journal groups versus 13 independent citing journals, the actual transmission is 726.75 mbits in 1984 and 732.23 mbits in 1985.[xii] The prediction on the basis of 1984, is 731.81 for 1985, i.e., the prediction now covers 92.3% of the 5.48 mbits increase in the transmission (as against 67.1% in the previous case).

If we subsequently assume that the citing action is not independent but completely grouped into the same three groups as the cited structure, the transmission is 669.33 mbits in 1984, and 672.43 in 1985. Now the prediction on the basis of 1984 is 673.02 mbits, which is 19.0% more than the observed increase in the transmission in 1985. Obviously, the assumption of complete grouping on both the cited and the citing side overestimates the structural coupling.[xiii]

In summary: I have elaborated here only an elementary model as an example. It is crude, among other things, since I did not allow for more groupings than the one into three groups, and I assumed that the one-year difference was an adequate time-scale. However, within this model a fit larger than 0.92 is obtained if we assume that the cited side is structured, and that on the citing side the journals behave independently.

## Consequences for the history of the coupled systems

The improvement in the prediction by action (formula (13)) is necessarily positive, since it is equal to the message to the system that action has taken place (i.e., formula (9) above); and this latter formula can be shown to be necessarily positive (Theil 1972, pp. 59f). Therefore, the prediction of "structure given action" at the later stage is always improved if we know how structure conditioned action at the previous stage. In each cycle, there is an increase of expected information content, since the new (larger) value $H_{(A|B)}$ will be the initial value ($H_{(A)}$) for the next cycle. Therefore, a structure/action contingency produces probabilistic entropy, and thus makes possible a history.[xiv]

Let me focus here on what it means theoretically and formally to say that structure gains probabilistic entropy by action. Intuitively it means that structure becomes more uncertain by action. Action incessantly adds to the uncertainty which prevails in the network. One can reduce this uncertainty only by the introduction of redundancy. (I return to this latter option in a later section.)

How does the generation of uncertainty work? Let us decompose the *a posteriori* information content into the *a priori* ones, and see what is added. After a given event, the total uncertainty in the structure can be written as follows:

$$H_{(A|B)} = - \Sigma\, q_{(A|B)} * \log(q_{(A|B)})$$

By using Bayes' formula, we can evaluate this *a posteriori* result into its *a priori* components, as follows:

$$H_{(A|B)} = - \Sigma\, \frac{p_{(A)} * p_{(B|A)}}{p_{(B)}} * \log\left\{\frac{p_{(A)} * p_{(B|A)}}{p_{(B)}}\right\}$$

$$= - \Sigma\, [p_{(A)} * \{p_{(B|A)} / p_{(B)}\}] * [\log\{p_{(A)}\} + \log\{p_{(B|A)} / p_{(B)}\}]$$

(See a previous section for the interpretation of $\{p_{(B|A)} / p_B\}$ as an *a priori* system.)

$$H_{(A|B)} = - \Sigma\, p_{(A)} * \log\{p_{(A)}\} - \Sigma\, \{p_{(B|A)} / p_{(B)}\} * \log\{p_{(B|A)} / p_{(B)}\} +$$
$$- \Sigma\, p_{(A)} * \log\{p_{(B|A)} / p_{(B)}\} - \Sigma\, \{p_{(B|A)} / p_{(B)}\} * \log\{p_{(A)}\}$$

$$= H_{(A)} + H_{(B|A)/(B)} +$$
$$- \Sigma\, p_{(A)} * \log\{p_{(B|A)} / p_{(B)}\} - \Sigma\, \{p_{(B|A)} / p_{(B)}\} * \log\{p_{(A)}\}$$

$$= H_{(A)} + H_{(B|A)/(B)} +$$
$$- \Sigma\, p_{(A)} * \log\{q_{(A|B)} / p_{(A)}\}$$

$$- \Sigma \{p_{(B|A)}/ p_{(B)}\} * \log[q_{(A|B)}/ \{p_{(B|A)}/ p_{(B)}\}]$$

$$= H_{(A)} + H_{(B|A)/(B)} +$$
$$+ \Sigma p_{(A)} * \log\{p_{(A)}/ q_{(A|B)}\}$$
$$+ \Sigma \{p_{(B|A)}/ p_{(B)}\} * \log[\{p_{(B|A)}/ p_{(B)}\} / q_{(A|B)}] \qquad (17)$$

Thus, the total uncertainty of the system *a posteriori* is equal to the sum of the uncertainties of two *a priori* systems (*A* and *(B|A)/B*) plus the sum of the information values of the messages that these systems have merged into one *a posteriori* structure. The sum of these two additional terms[xv] is equivalent to what may also be called the "in-between" group uncertainty ($H_0$) upon decomposition of the total uncertainty in $H_{(A)}$ and $H_{(B|A)/(B)}$. Note the analogy between "later" and "more aggregated": both contain more uncertainty. However, this "in-between group" uncertainty is composed of two terms: the difference which it makes for the one *a priori* subset in relation to the *a posteriori* set, and the difference it makes for the other. An update cycle affects two (or more) *a priori* systems asymmetrically!

In the above formula, I decomposed the *a posteriori* expected information content $H_{(A|B)}$ into its various parts, which were given a meaning (on the right-hand side of the equation) in terms of the *a priori* states of the respective systems. Let me note that this is paradoxically what Bayesians always do, although they use a different rhetoric. The Bayesian frame of reference is not the *a posteriori* situation, but the *a priori* one. For example, the Bayesian philosopher asks what it means for the prior hypothesis that a piece of evidence becomes available (see, e.g.,

Howson and Urbach 1989). That the hypothesis (or, analogously, the subjective belief) itself may have changed, and thus no longer be the same hypothesis, is for him/her usually of little concern. The Bayesian, however, is not interested in the further development of the *a priori* stage into the *a posteriori* one thanks to the new evidence, but only in the corroboration or falsification of the *a priori* hypothesis (see also: Leydesdorff 1992b). From a social science perspective, however, one is interested in what happened empirically, and not only in what this means in terms of the previous stage. However, explication of the meaning of what happened in terms of what was *a priori* adds obviously to the redundancy, and therefore has a positive function for the reflexive understanding. Note that some authors have wished to reserve the term "information" for denoting this (human) neg-entropy (e.g., Brillouin 1962; Bailey 1990), in contrast to probabilistic entropy or noise.

Redundancy in social systems

It is counter-intuitive that the structure/action contingency would generate only uncertainty. "Structure" is usually believed to reduce uncertainty for "action". However, because of the equation between uncertainty, entropy and information in the Shannon/Luhmann paradigm, we must also redefine what is meant precisely by this reduction of uncertainty by structure (or action) in social systems.

In general, reduction of uncertainty is possible only when (Shannon-type) information is relatively annihilated by an increase in redundancy. Redundancy is defined as the complement of the expected information content of a system to its maximal information content. Since necessarily $H_{(B|A)}$ <=

$H_{(B)}$, the conditioning of the actors by structure in action *reduces* uncertainty. However, as shown in the previous section, action subsequently adds to the uncertainty that prevails in a structure.

Until now, we have largely concentrated on the model of one action system *B* in relation to structure *A*. However, all other action systems (*C*, *D*, etc.) conditionalize structure as well. By using the same equation, it will be clear that these conditionalizations lead to a further reduction of the uncertainty in the structural system. Whether this will have an effect on the action system *B* depends on whether the redundancy is generated in that part of the information content of structure *A* which is transmitted to this action system (i.e., $H_{(A)} - H_{(A|B)}$) or in the remainder of the uncertainty in *A* (i.e., $H_{(A|B)}$). In the latter case, this redundancy will have no effect on *B* as an actor, while in the former case the reduction of the uncertainty for actor *B* by structure *A* (i.e., $H_{(B)} - H_{(B|A)}$) in the action will be *smaller* than without the redundancy by the other actor. Therefore, $H_{(B)}$ and $H_{(B|A)}$ will be more equal; the log-factor in the algorithm (i.e., formula (13) above) will be closer to zero, and the impact of action upon structure will thus also be smaller.

In summary, the redundancy brought about by other actors may either be irrelevant for each single actor or may reduce the determination by the system, i.e., may free the individual actor more from the system; but by the same token then it also reduces the impact of his/her actions on the system.

Let me demonstrate this with the example of a family as a social system and of two parents and two children as actors. The maximal information content of this family system is $^2\log(4) = 2$ bits. Empirically, family life is usually structured, and thus the actual information content of the system will be lower.

Now, let us begin with an action by one of the neighbours. This cannot affect this family system directly, since the neighbour is not a member of it: it can only affect the system if it either induces action by one of the members of the family for reasons which lie within the context of this system, i.e., at the higher level of how this system is integrated in, e.g., the neighbourhood-system. The action by a neighbour generates redundancy in the neighbourhood-system, and thereby reduces the overall uncertainty in which the family-system has to operate. This redundancy may be partly transmitted to individual members of the family to varying degrees. One may wish to call this "social control".

Note that at the neighbourhood level we can either attribute transmissions directly to individual actors or use grouping rules (by using formula (6) above). In this case, some uncertainty can also be attributed to the "in between group" uncertainty.

Within the family system, the children can, for example, argue with each other. This as an action adds to the entropy of this system. Each parent can try to intervene in this argument, and may successfully change the course of action of each of the children. If so, this parent reduces uncertainty for the children by using his/her impact on the family structure in order to conditionalize this system so that it affects the child's room for action. However, at the same time, this action also necessarily contributes to the overall uncertainty in the family system. (The restriction may, for example, be ambiguous.) The paradoxical consequence is that the parent can only reduce the child's room for action by adding to the uncertainty of the family system, and by thus giving the child more freedom as an actor, i.e., reducing its determination by the system, and thereby reducing its expected impact on the system.

However, if the parents go away, they may take all kind of decisions which then heavily influence the further development of this family system. However, these actions cannot affect the child's room for action at the same moment. First, the family system has to be updated with the information content of the message of these parental actions. The effect of the update may affect the room for the children to argue with each other or not. For example, the decision to add another element to the system, i.e., to take on a third child, undoubtedely enlarges the future possibilities to argue, since it extends the maximal entropy of the system.

The need for an update before a change in the relations can be achieved points to another important mechanism of reduction of uncertainty in social systems: *a priori* structures hold as long as they are not updated. In the meantime a lot of entropy may have been produced within the system, but this new information cannot intervene in the relation of each individual actor to the system before the update. Since the individual actor is conditionalized by the prior state of the system, (s)he has to cope with less uncertainty than if (s)he had to take all available information into account. Thus, information may even have been destroyed (e.g., as a result of the generation of redundancy by another system) before the system is updated.

This mechanism is particularly useful for an understanding of the working of a functionally differentiated system. In this case, the total information content of the social system is disaggregated in accordance with specific grouping rules, with the consequence that an actor in one of the subsystems is confronted only with that part of the uncertainty which is of special relevance to him/her in this subsystem. The subsystem is provisionally shielded against perturbations in the environment. However, from time to time the system has to be updated. Then, the disaggregation may change correspondingly, and the actor will be conditionalized by a new set of structural arrangements.

In my opinion, along these lines we can also gain a better understanding of the meaning of Giddens' (1979) notion of "the duality of structure". Structure, according to Giddens (1984), is a "set of rules and resources" which is recursively implicated in action. However, Giddens acknowledges that in society there are various "structures" which are therefore defined as "rule-resource sets." Obviously, each of them implies a specific grouping of the overall uncertainty. This leads to the reduction of uncertainty for social action within each subset. Institutionalization and routines can serve as means for the imposition of this redundancy.

However, after the action(s), i.e., *a posteriori*, the system and the relevant subsystems have gained in entropy. Therefore, the decomposition may also have to change. However, the shielding now works as a delay. *A posteriori* there is necessarily more uncertainty in comparison with the situation in which the *a priori* demarcations were made, but the system does not have necessarily to update. Giddens' problem of the "duality of structure" can now be considered as the problem that the system has no yardstick to evaluate this change other than its prior organization, while, as noted in the previous section, the additional uncertainty cannot be fully explained in terms of the *a priori* uncertainty, since *a posteriori* it includes the information content of the message that each system has gone into the *a posteriori* state. Since these messages are asymmetrical for each system, the system cannot oversee how much uncertainty is generated overall, but only that part to which it can recur. It therefore has to decide whether it wishes to maintain the previous organization of its identity under the risk of not incorporating as much uncertainty as it might be able to contain after an update, or risk losing its (recursive) identity. In summary, a central question for social and evolutionary systems is whether or not it is timely for an update.

The continuous creation of uncertainty in its contingent relation requires counterbalancing mechanisms to stabilize each system. The reduction of uncertainty by the introduction of redundancy, reflexively reinforced by giving specific parts of the *a priori* information content symbolic meaning, plays a crucial role here. However, the issue of how social and psychological stability are maintained under pressure from the continuous relevance of events in structurally coupled systems goes beyond the limits of this study.

### Redundancy on the side of actors

Similarly, human actors are self-referential systems which have to integrate in one way or another the information contained in the message of any event which is a result of their interaction with a relevant social structure. Again, the uncertainty can only be integrated after proper normalization, and in terms of the system itself. In order to understand the communication, the uncertainty has to be processed internally by giving it a meaning in terms of the actor system. To give meaning is therefore a reflexive action, i.e., it is an action internal to the actor system. In this action, the incoming uncertainty is again evaluated against preset and recursive standards, i.e., the information content of the primary observation is observed in a second-order cybernetics. However, in this case the organization is not a grouping of the total uncertainty in terms of a differentiation, as in the case of society, but in terms of the reflexive identity of the actor(s).

How actors reflexively give meaning to incoming uncertainty, is in itself the subject of psychology (cf. Luhmann 1984). However, the mechanism of balancing between redundancy and variation which I described above for social systems must have its cybernetic analogon in the action system, since it has survival value. The structure/action contingency necessarily generates uncertainty; and both of the systems involved in this contingency relation have to provide mechanisms for selection and stabilization. Otherwise, they would rapidly disintegrate (i.e., become chaotic), and thus no longer be able to reproduce themselves as systems.

Conclusion

The algorithm which I specified in this article only teaches us about the uncertainty which becomes available through interactions in the contingent relations between "structure" and "action". How this interaction is integrated can only be answered within the framework of respectively a theory of communication, i.e., a sociology, and a theory of the actor, i.e., a psychology. There is still a great difference beteen a single general algorithm for structure/action contingencies and a general theory of social action.

However, the derivation of this one algorithm and the discussion of its various consequences for the interacting systems elucidates the power of the newly emerging cybernetic paradigm in sociology. The implied "Gestalt switch" consists of two central elements: Shannon's (1948) equation of information with *un*certainty (see, e.g., Hayles 1990), and Luhmann's (1984) understanding of society as an "autopoietic" communication system with its own forms of organization (e.g., functional differentiation), and which is analytically different from (but structurally coupled with) the organization of individual actors as autopoietic systems (cf. Leydesdorff 1993b). One purpose of this study has been to show that the implied problems can consistently be brought down to the level of puzzles which can be solved.

Table 1

13 journals used for the construction of an aggregated journal-journal citation network

| journals: | *grouping:* |
|---|---|
| Chemical Physics | *chemical physics* |
| Chemical Physics Letters | *chemical physics* |
| Inorganic Chemistry | *inorganic chemistry* |
| J. of the American Chemical Society | *organic chemistry* |
| J. of Chemical Physics | *chemical physics* |
| J. of the Chemical Society- Dalton Transactions | *inorganic chemistry* |
| J. of Organic Chemistry | *organic chemistry* |
| J. of Organometallic Chemistry | *inorganic chemistry* |
| J. of Physical Chemistry | *chemical physics* |
| Molecular Physics | *chemical physics* |
| Physical Review A | *chemical physics* |
| Tetrahedron | *organic chemistry* |
| Tetrahedron Letters | *organic chemistry* |

\*. A previous version of this paper was presented at the Joint EC/Leiden Conference on Science and Technology Indicators 1991 (cf. Leydesdorff 1993a).

i. I use the binary base of the logarithm throughout this study, and therefore express the information in bits.

ii. Strictly speaking, chi-square tests only independence; it provides little information about the strength or form of the association between the two variables.

iii. The so-called likelihood ratio chi-square ($L^2 = 2 \Sigma_i \Sigma_j F_{ij} \ln (F_{ij} / \hat{F}_{ij})$) is equally decomposable into interpretable parts that add up to the total. This measure is in essence an information-theoretical formulation of the chi-square. See also: Krippendorff (1986).

iv. For a proof see Theil (1972, pp. 59f.).

v. In the case of $q = p$, no information is added or lost; since the log(1) = 0, $I$ vanishes. Note that a zero in the prior distribution would make a non-zero value in the posterior distribution a complete surprise, and therefore, $I \rightarrow 4$. See also: Leydesdorff (1990 and 1992a).

vi. For a further elaboration of the relations among statistical decomposition analysis, regression analysis, and Markov chain analysis, the reader is referred to Theil (1972).

vii. Even if action fails to take place this may also condition the network. As we will see below, structure normalizes action, and therefore contains an expectation value. Note also that action may lead to a null-message, i.e., to a message containing no information. See, e.g., Bertsekas and Tsitisklis 1989.

viii. In Bayesian philosophy, this term is a normalization constant because of the logical complementarity of the hypothesis and its negation (see also: Pearl 1988, p. 32.)

ix. In order to minimize the amount of expected information content originated by missing values in otherwise similar distributions, I have replaced all missing values in this study with the value of five, since this is the cutoff level of the printed edition of the <u>Journal Citation Reports</u> of the <u>Science Citation Index</u> from which the data were obtained (Garfield 1972). For a further discussion of this matrix see (Leydesdorff 1991).

x. Note that we can also use formula (13) to compute the transmission in a static model.

xi. The same data were analyzed in more detail in (Leydesdorff 1991).

xii. These relatively smaller transmissions in absolute terms are larger parts of the total uncertainty in the respective matrices, since the matrices are differently shaped, and can therefore hold less entropy.

xiii. The assumption of grouping in only the citing action overestimates the coupling with 5.9%.

xiv. However, having a history does not imply that this history is always important for later developments. For example, the system may have the Markov property or go through path-dependent transitions. In such cases, historical information can lose its relevance for further developments (cf. Leydesdorff 1992a).

xv. Since these are information contents of messages about change, they can be shown to be necessarily positive (see, Theil 1972, pp. 59f.).